\documentclass[letterpaper, 10 pt, conference]{ieeeconf}  

\IEEEoverridecommandlockouts                              

\usepackage{graphics} 
\usepackage{graphicx}
\usepackage{amsmath} 
\usepackage{capt-of,etoolbox}
\usepackage{amsfonts}
\usepackage{listings}
\usepackage{xspace}
\usepackage{multirow}
\usepackage{array}
\usepackage[table,xcdraw]{xcolor}
\usepackage{url}

\usepackage{mathtools}

\DeclarePairedDelimiter\floor{\lfloor}{\rfloor}

\DeclareGraphicsExtensions{.pdf,.png,.jpg}

\newcommand{\algoname}{SkiMap}

\begin{document}

\makeatletter
\patchcmd{\@maketitle}{\end{center}}{{\myfigure{}\par}\end{center}}{}{}
\makeatother

\newcommand\myfigure{%
\centering
\vspace{0.2cm}
    \includegraphics[width=0.90\linewidth]{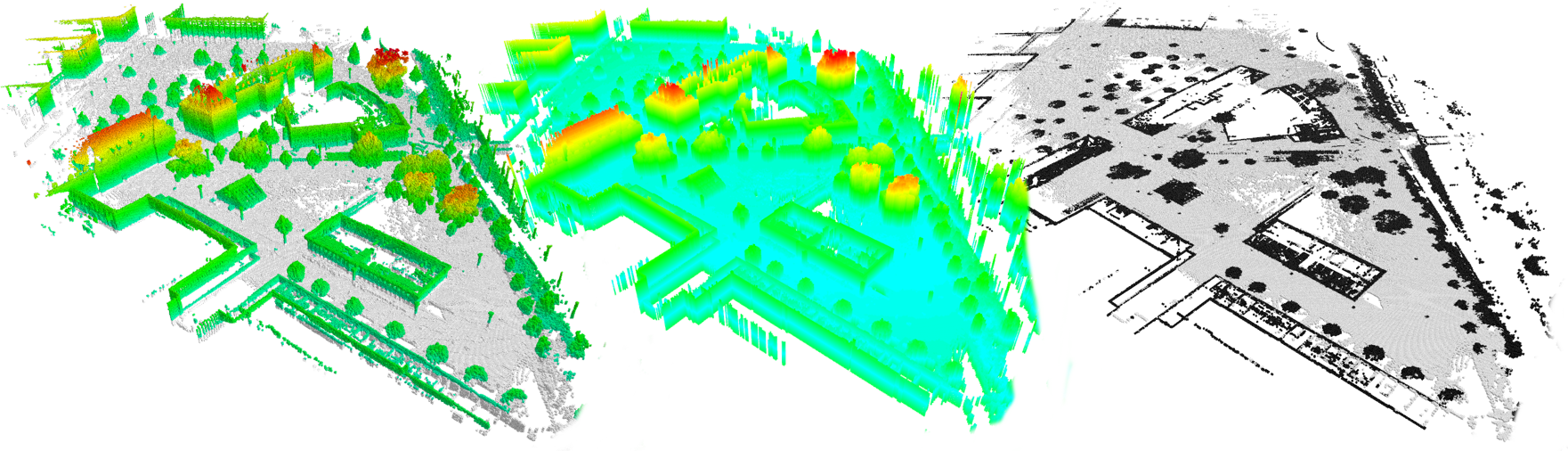}
\captionof{figure}{\algoname{} encodes seamlessly a full 3D reconstruction of the environment (left), a height map (center) and a 2D occupancy grid (right). The three representations can be delivered on-line with decreasing time complexity. The displayed maps have been obtained on the \emph{Freiburg Campus} dataset.}
\label{fig:teaser}

}

\title{\LARGE \bf 
 \algoname{}: An Efficient Mapping Framework for Robot Navigation
}

\author{Daniele De Gregorio$^{1}$ and Luigi Di Stefano$^{1}$
\thanks{$^{1}$DISI, University of Bologna, Italy}%
}

\maketitle

\thispagestyle{empty}
\pagestyle{empty}

\begin{abstract}

We present a novel mapping framework for robot navigation which features a multi-level querying system capable to obtain rapidly representations as diverse as a 3D voxel grid, a 2.5D height map and a 2D occupancy grid. These are inherently embedded into a memory and time efficient core data structure organized as a \emph{Tree of SkipLists}. Compared to the well-known \emph{Octree} representation, our approach exhibits a better time efficiency, thanks to its simple and highly parallelizable computational structure, and  a similar memory footprint when mapping large workspaces. Peculiarly within the realm of mapping for robot navigation, our framework supports real-time erosion and re-integration of measurements upon reception of optimized poses from the sensor tracker, so as to improve continuously the accuracy of the map.
\end{abstract}

\section{INTRODUCTION}

Key to autonomous robot navigation is the ability to attain a sufficiently rich perception of the environment, so to allow the robot to plan a path, localize itself and avoid obstacles. This kind of perception is  realized through suitable sensors and algorithms. As for the former, laser rangefinders have  traditionally been employed to capture a planar view of the surroundings, while visual sensors, and in particular RGB-D cameras, are becoming more and more widespread on account of their potential to model the environment in 3D. 
In the space of algorithms, many proposals concern full-fledged SLAM (Simultaneous Localization And Mapping) systems aimed at both building a map of the workspace and localizing the sensor (i.e. the robot) therein. Other works, differently, are focused on the mapping task and address issues such as memory efficiency, quite mandatory to enable navigation in large spaces, and time efficiency, which concerns creating on-line the  representation required by the navigation system, such as a 2D occupancy grid to plan a path through the environment or a 3D reconstruction to avoid obstacles reliably while a robot moves around or a 2.5 (aka height) map to assess free space at the flight altitude  of a MAV (Micro Aerial Vehicle). Along the latter research line, in this paper we focus on mapping and present a novel approach, dubbed \algoname, which is particularly time efficient and flexible enough to support seamlessly different kinds of representations that may be delivered on-line according to the application requirements (Figure \ref{fig:teaser}).  

Another favorable trait of our mapping framework is the ability to erode and fuse back measurements in real-time upon receiving optimized poses from the sensor localization module in order to improve the accuracy of the map. Indeed, many recent sensor localization algorithms  based on visual data can perform pose optimization on-line, e.g. upon detection of a loop closure, which holds the potential to continuously improve the map as long as sensor measurements may be injected therein according to the new optimized poses rather than the old ones. 

Our framework has been implemented as a ready-to-use ROS \cite{ROS09} package freely distributed for research and education purposes\footnote{\url{https://github.com/m4nh/skimap_ros}}. The package can be configured either to achieve mapping in conjunction with any external sensor localization  module or as a full-fledged SLAM system, \emph{Slamdunk} \cite{Fioraio2015} providing camera poses optimized on-line in the latter option.

\section{RELATED WORK}
As described in \cite{Thrun05}, the classical mapping approach for robot navigation is the 2D occupancy grid.  Accordingly, sensors measurements (typically from planar laser scanners) are fused into a 2D Grid wherein each tile (i.e. a square chunk of the space) contains an occupancy probability which can be interpreted as the likelihood that the tile belongs to an obstacle. Many robot navigation systems, often referred to as \emph{Grid-Based SLAM}, rely on this 2D occupancy grid  \cite{Grisetti07}, which is available in ROS \cite{ROS09} and can be considered as a baseline for robot navigation.
\par 
Yet, planar sensing and related 2D mapping may not be reliable enough due to defective reconstruction of the environment, e.g. when dealing with a MAV (micro aerial vehicle) for indoor navigation, or, more generally with any Mobile Robot that cannot be modeled as a bi-dimensional agent or any robot that has more than 3 DOFs. A conceptually straightforward approach to pursue 3D mapping when deploying sensors, e,g. visual sensors, capable of delivering 3D measurements would consist in extending the occupancy map to a 3D Grid by cutting the 3D space into \emph{Voxels} (i.e. small cubes) \cite{Tabak89}, each voxel storing the probability for an obstacle to be located therein. However, handling a 3D occupancy grid may easily become impractical when dealing with large workspaces due to the excessive memory footprint; for example, should the probability stored in each voxel be encoded as a float number (\emph{4 Bytes}), the 3D occupancy grid would require as many bytes of memory as

\small
\begin{equation}\label{memory_full}
  M_{OG} = \frac{x \times y \times z}{r^3} \times 4  
\end{equation}
\normalsize

$x,y,z$ being the sizes of each of the dimensions of the workspace and $r$ the voxel resolution (i.e. the voxel size expressed in the same units as $x,y,z$).

A quite popular memory efficient alternative to the 3D occupancy grid is the Octree \cite{meagher82}, whereby the 3D Space is recursively partitioned into octants (octants being equivalent of quadrants in the 3D space) until voxels take the desired resolution. As such, this data structure is a \emph{tree} in which each node has exactly 8 children; unlike the 3D grid, though, the Octree avoids modeling the empty space as only leaf and inner nodes associated with occupied space need to be allocated, thereby yielding significant memory savings when representing large environments. Hence, well-known mapping frameworks like Octomap \cite{Hornung2013} rely on this kind of data structure to build the required workspace representation. In particular, Hornung \emph{et. al.}\cite{Hornung2013} shape their data structure so that voxels (leafs of the Octree) are the only nodes storing mapping information, all other ones containing references to children only. Therefore, the memory occupancy in bytes can be expressed as:

\small
\begin{equation}\label{memory_octree}
  M_{OCT} = n_{leaf} \times B_{leaf} + n_{inner} \times B_{inner}
\end{equation}
\normalsize

where $B_{leafs}$ and $B_{inner}$ are the occupancy in bytes of \emph{leaf} and \emph{inner nodes}, and $n_{leaf}$,$n_{inner}$ the number of \emph{leaf} and \emph{inner nodes} corresponding to non-empty space, respectively. Accordingly, the memory footprint is dependent on the amount of space actually occupied  and not on the overall size of the environment. 

However, using an Octree rather than a 3D grid implies a space vs. time trade-off, the memory footprint is reduced at the expense of the query time, the computational complexity of a random voxel access being just $\mathcal{O}(1)$ for a 3D Grid, as large as $\mathcal{O}(\log d)$ for an Octree ($d$ denoting the depth of the tree). This issue has motivated a recent proposal by Labschutz \emph{et. al.}\cite{Labschutz16} who mix the two approaches into a novel data structure referred to as \emph{Jittree} and fully managed by the GPU.\par


Researchers have also explored other solutions, such as  \emph{Multi-Level Surface Maps} \cite{Triebel06} and \emph{Multi-Volume Occupancy} \cite{dryanovski10}, aimed at ameliorating the memory efficiency of the data structures, referred to as height or 2.5D maps, that endow a 2D Grid with measurements concerning the height of obstacles. In particular, information about occupied and free-space is accounted through  
  a dynamic 2D grid where each element is a list of voxels. Thus, akin to   Eq.\ref{memory_octree}, the memory occupancy may be expressed as:

\small
\begin{equation}\label{memory_mls}
  M_{MLS} = \frac{(x \times y )}{r^2} \times B_{tile} + n_{voxels} \times B_{voxel}
\end{equation}
\normalsize

where $x,y$ are the sizes of the projection of the workspace onto a plane (e.g. the ground plane), $r$ the resolution of the above mentioned  2D grid, $B_{tile}$ is the occupancy in bytes of a tile of the grid, $B_{voxel}$ that of each element of a voxel list, $n_{voxels}$ the number of voxels dealing with non-empty space. This kind of approach compares favorably w.r.t. the Octree in terms of memory footprint \cite{dryanovski10}, though, again, at the expense of time complexity: as the data structure is basically a linked-list on top of a grid, a random voxel access takes  $\mathcal{O}(n)$ ($n$ being the number of voxels in a list). 

As already mentioned, a basic 2D Map built from planar range sensors is ofter not reliable for navigation due to lack of information concerning the height of obstacles. On the other hand, due to the complexity of pursuing path planning directly in the 3D space, most proposals, such as \cite{Gutmann05}\cite{Maier12}\cite{Biswas12}, deploy the rich information embedded into a 3D map so to create a reliable 2D projection that is actually used for the sake of planning. Consequently, the time efficiency of visiting both the entire 3D Map as well as the local neighborhood of a given 3D point (aka radius search) are crucial aspects in the selection of the right data structure.

The mapping framework proposed in this paper, dubbed \algoname, features a memory footprint similar to an Octree when dealing with large environments and a better time complexity, i.e. $\mathcal{O}(\frac{\log n }{k})$, thanks to a highly-parallelizable computational structure. Besides, it inherently embodies  a 3D, 2.5D and 2D map that can be delivered with a time complexity which decreases alongside the richness of the representation. Moreover, unlike the previously mentioned mapping frameworks proposed for robot navigation, \algoname{} can deploy the ability of the sensor localization  module to deliver optimized poses in order to update the map in real-time. Indeed, such a task is pursued nowadays only by state-of-the-art SLAM systems, such as \cite{Fioraio20152} or \cite{Dai2016}, aimed at producing high-quality 3D scans of the workspace and meant to run on desktop computers equipped with high-performance and power-hungry GPU cards, an application domain quite different from mobile robotics which calls for compact, low-power computing platforms mounted on-board. In particular, the recent proposal in \cite{Dai2016} represents the 3D Space using \emph{VoxelHashing}  \cite{niessner2013hashing}, which is a memory-efficient data structure managed by the GPU which enables fast random voxel access, but turns out to be vastly inefficient for radius search, which is a key requirement for robot navigation.


\section{\algoname{}  MAPPING ALGORITHM}
 
In this section we explore the \algoname{} algorithm in its entirety, describing the key data structure as well as how to carry out mapping differently from standard approaches like  \emph{Octree} or 3D Grid. Furthermore, we  highlight the inherent parallelism of the proposed data structure, which is conducive to notably improved time efficiency  in key tasks dealing with robot navigation. 
 
\subsection{DATA STRUCTURE: TREE OF SKIPLISTS}\label{section_data_structure}
\begin{figure}[thpb]
      \centering
      \includegraphics[width=0.35\textwidth]{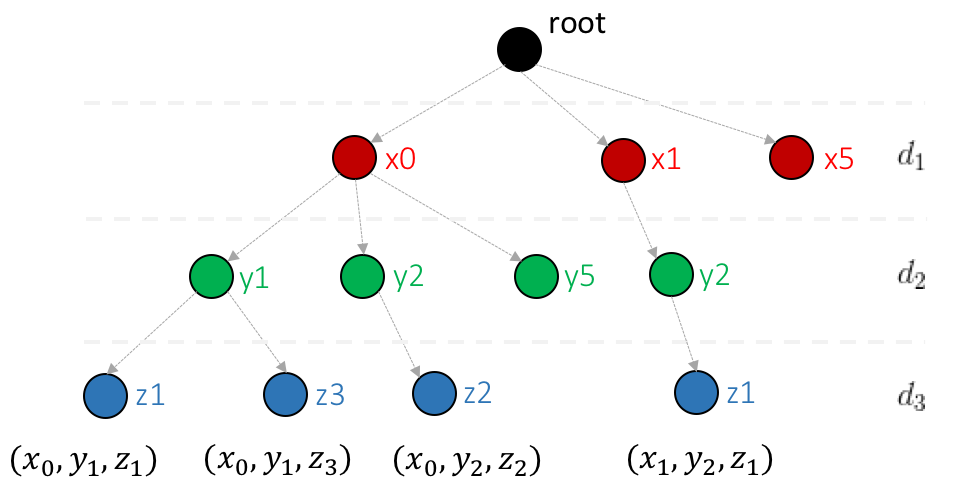}
      \caption{Tree structure to group voxels according to their coordinates. The maximum depth of the tree is \emph{3},  nodes with depth $d_3$ being voxels while those with depths $d_1,d_2$ being transient nodes. Nodes at depths $d_1,d_2$ store only integer numbers representing the associated quantized coordinate, while voxels (\emph{blue nodes}) can be deployed to store user data, such as for example Occupancy Probability \cite{Thrun05}.  }
      \label{fig:skiplist_tree}
   \end{figure}
   
\begin{figure}[thpb]
      \centering
      \includegraphics[width=0.4\textwidth]{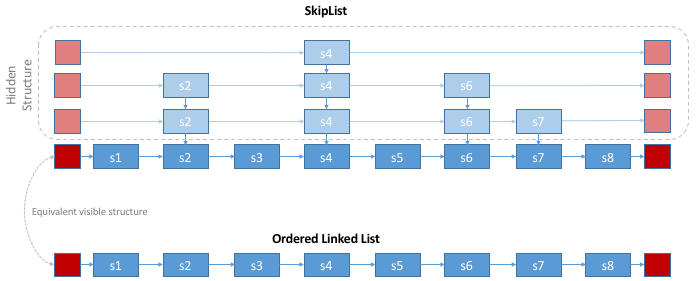}
      \caption{The visible part of a \textbf{SkipList} is identical to a \textbf{LinkedList}. The hidden segment of a SkipList shall ensure a random access complexity of $\mathcal{O}(\log n)$ rather than  $\mathcal{O}(n)$.  }
      \label{skiplist_vs_linkedlist}
   \end{figure}
   
   \begin{figure}[thpb]
      \centering
      \includegraphics[width=0.35\textwidth]{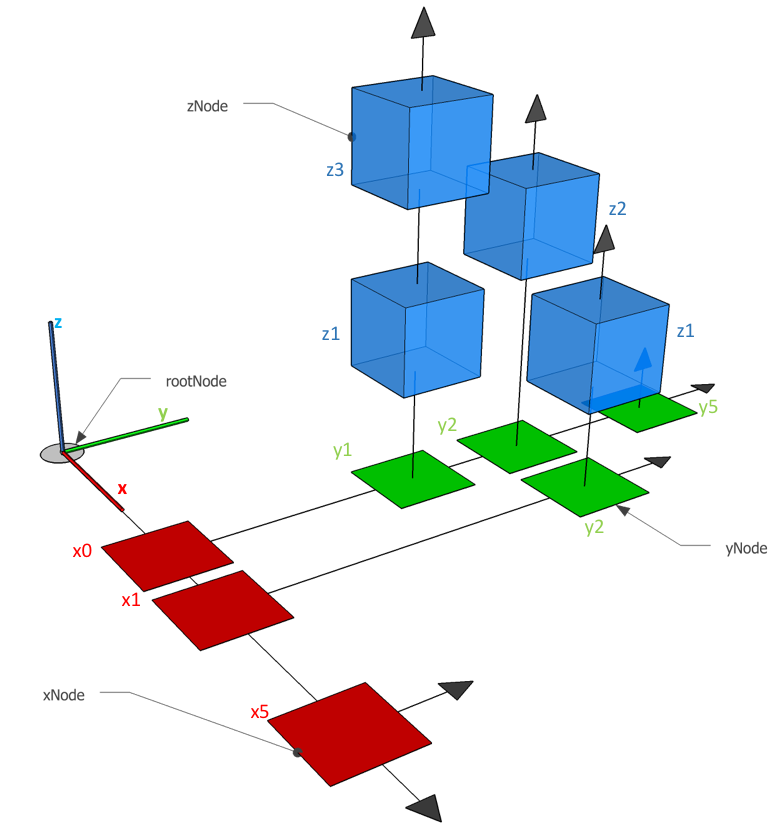}
      \caption{Grouping voxels into a \textbf{Tree of SkipLists}.  Each voxel (\emph{blue box}) is linked to the \emph{rootNode} by a \emph{yNode} (green tile) which in turn is linked to a \emph{xNode} (red tile). }
      \label{skiplist_volume}
      
   \end{figure}

\algoname{}   relies on the basic concept of grouping voxels within a tree as outlined in Figure \ref{fig:skiplist_tree}. The actual voxels are nodes at depth \emph{3}, which are grouped into nodes at depth \emph{2} according to equal quantized $(x,y)$ coordinates, the nodes at depth \emph{2} in turn grouped into nodes at depth \emph{1} according to equal quantized $x$ coordinates.  However, adopting a classical tree  structure to realize the concept illustrated in Figure \ref{fig:skiplist_tree} would not be efficient because of the unbounded number of siblings at each depth level (unlike the \emph{octree}, in turn, where each node has always \emph{8} children). Indeed, should the children of each node be stored in a ordinary list, performing a random access would exhibit $\mathcal{O}(n)$ complexity. To overcome this efficiency issue, we adopted a rather uncommon data structure called \emph{SkipList} and proposed by Pugh \emph{et. al.} \cite{Pugh90}. As shown in Figure \ref{skiplist_vs_linkedlist} a \emph{SkipList} is apparently similar to an \emph{Ordered Linked List}, but the former also stashes a super-structure aimed at bringing the computational complexity associated with random access from $\mathcal{O}(n)$ down to $\mathcal{O}(\log n)$. In a \emph{SkipList} elements are kept ordered, and thus, compared to an ordinary list, insertion time  grows from  $\mathcal{O}(1)$ to $\mathcal{O}(\log n)$ due to each insertion requiring a search.

Figure \ref{skiplist_volume} shows the actual realization of the concept illustrated in Figure \ref{fig:skiplist_tree}. A  first \emph{SkipList} keeps track of quantized \emph{x} coordinates, thereby realizing depth level \emph{1} of Figure \ref{fig:skiplist_tree}; the items of the first \emph{SkipList} are referred to as \emph{xNodes} and colored in red in Figure \ref{skiplist_volume}; each \emph{xNode} is in turn a \emph{SkipList} which keeps track of quantized \emph{y} coordinates, thus implementing depth level \emph{2} of Figure \ref{fig:skiplist_tree}; the items in these nested \emph{SkipLists} are dubbed \emph{yNodes} (green) in Figure \ref{skiplist_volume}; eventually, each \emph{yNode} is a \emph{SkipList} of \emph{zNodes} (blue), which represent the actual voxels and provide the containers for any kind of user data. Therefore, the concept shown in Figure \ref{fig:skiplist_tree} is realized by a novel data structure that may be thought of as a \textbf{Tree of SkipLists}. 
It is worth pointing out that with the proposed data structure the coordinates of a voxel can be obtained by iterating through its predecessors and thus need not to be stored in the containers together with user data; for example for the voxel referred to as $z_3$ in Figure \ref{skiplist_volume}, iterating back through predecessors provides  coordinates  $(x_0,y_1,z_3)$. A similar technique is used in \emph{Octomap} to avoid coordinates storage in the leaves of the octree \cite{Hornung2013}. \par

As a detailed description of the \emph{SkipList} is outside the scope of this paper and can be found in \cite{Pugh90}, we conclude this section with a brief review of the key concepts related to this topic. A \emph{SkipList} is a multi-level linked-list in which the first level is a list containing all the elements ordered by a \emph{Key} (each node being a pair $<Key,Value>$) level $i$ contains about half elements of level $i-1$, still ordered by \emph{Key}. Similarly to a \emph{Binary Tree}, a search is performed starting from level $i=i_{max}$ down to $i=1$ in $\mathcal{O}(\log n)$, at the expense of memory footprint (due to replicated elements). \par
Depth (i.e. number of levels) is a parameter of the \emph{SkipList} to be chosen based on application settings. Indeed, there exists an upper limit beyond which one gets no further benefits in terms of timing performance while significantly increasing memory footprint. As reported in Table \ref{table-skiplist-depth-comparison}, this is vouched also by our experimental findings. 

\begin{table}[]
\vspace{0.25cm}
\centering
\resizebox{0.4\textwidth}{!}{
\begin{tabular}{|l|l|l|l|}
\hline
\textbf{SkipList Depth}  & \textbf{ Integration Time}        & \textbf{Visiting Time}   & \textbf{Memory}     \\ \hline
4                        & 56 ms                                 & 215 ms                        & \textbf{432 MB}               \\ \hline
\rowcolor[HTML]{DDDDDD} 
{\color[HTML]{333333} 8} & {\color[HTML]{333333} \textbf{29 ms}} & {\color[HTML]{333333} 269 ms} & {\color[HTML]{333333} 588 MB} \\ \hline
16                       & \textbf{29 ms}                        & \textbf{215 ms}               & 900 MB                        \\ \hline
32                       & 32 ms                                 & 259 ms                        & 1524 MB                       \\ \hline
64                       & 33 ms                                 & 258 ms                        & 2743 MB                       \\ \hline
\end{tabular}
}
\caption{Analysis of SkipList depth: tests performed on \textbf{Freiburg Campus} dataset with a resolution of $0.05m$. The Table reports the average computation time to integrate new sensor measurements ($\sim$180k points), the average time for a full visit of the map and the  memory footprint of the  map. \label{table-skiplist-depth-comparison} }

\end{table}

\subsection{VOXEL INDEXING}

As each node of our data structure is addressable by a \emph{Key}, we can use it to map real world coordinates to quantized indexes just as it would happen in a 3D Grid. Thus, to retrieve the voxel $\mathbf{v}(I_x,I_y,I_z)$ corresponding to a 3D point $\mathbf{p}(x,y,z)$:

\begin{equation}\label{eq_coordinates_conversion}
 I_x = \floor*{\frac{x}{r}},
  I_y = \floor*{\frac{y}{r}},
  I_z = \floor*{\frac{z}{r}}
\end{equation}

\noindent $r$ denoting, as usual, voxel resolution. Unlike a 3D Grid, however, we can use also negative indexes as they represent  \emph{Key}s of a map rather than simple indexes of an array. This is important for mapping applications as, more often than not, the ground reference of the map (aka \emph{Zero Reference Frame}) is not known a priori.
\par With our data structure, querying for a voxel $f(I_x,I_y,I_z)=\mathbf{v}$ consists in executing the iterative query $h(g(f(I_x),I_y),I_z)=\mathbf{v}$. Thus, with reference to Figure \ref{skiplist_volume}:

\begin{itemize}
\item $f(\bullet)$ retrieves a \emph{red tile} / \emph{xNode}
\item $g(\bullet)$ retrieves a \emph{green tile} / \emph{yNode}
\item $h(\bullet)$ retrieves a \emph{blue box} / \emph{zNode} / \emph{Voxel}
\end{itemize}

Each of three query function $f(\bullet),g(\bullet),h(\bullet)$ can result in either a \emph{Hit} or a \emph{Miss}. Moreover, each generic function $\phi(f(I_x))$ may be performed concurrently because it involves separate branches of the SkipList Tree (see again Figure \ref{skiplist_volume}). 

\subsection{PARALLELIZATION}

As highlighted in the previous section, the proposed data structure inherently provides for a high degree of parallelization. Besides, even a single \emph{SkipList} may enable a certain level of parallelization by using locks on nodes \cite{Pugh89C}. However, we decided to exploit only the high parallelism among voxel indexing operations enabled by our data structure while not deploying also the lock-based technique to further parallelize accessed within a \emph{SkipList}, mainly to maintain a lean and simpler code and secondly due to lock-based algorithm being often unpredictable, which makes them unsuited to real-time tasks. \par
As already mentioned the operations involving separated branches of the first level of our SkipList Tree, that is $f(I_{x_i}) \neq f(I_{x_j}) \rightarrow x_i \neq x_j $, can be performed in parallel, We can classify all the possible operations on the data structure into two main categories:

\begin{itemize}
\item
	\emph{Visiting Operations}: Visiting the whole tree (i.e. reaching each voxel of the map) consists in visiting all first-level nodes in parallel and collecting the results: 

\small
\begin{equation}
\pi(\Gamma) =  \sum_{i=min}^{max} \pi(f(I_{x_i}))
\end{equation}
\normalsize

\item 
	\emph{Updating Operation}: upon performing a generic update operation we cannot know in advance whether it will produce a new allocation or a deallocation or it will just update the content of an existing voxel without performing further search. To ensure that an update operation is not concurrent over others we can reuse again the previous technique: we assume that two update operations do not conflict if they belongs to two separate first level branches. In a typical application scenario we are given a set of sensor measurement to be integrated into the map, i.e. a set of 3D points: $C=\{\mathbf{p}(x,y,z)\}$.  Hence, we can group these points in subsets according to their first quantized coordinates
	
\small
\begin{equation}
C = \bigcup_{i=min}^{max} C_{x_i} \mid C_{x_i} : \{\floor*{\frac{x}{r}} = I_{x_i}\}
\end{equation}
\normalsize

so to perform the integration operation dealing with each of the subsets in parallel while ensuring no concurrent memory access. 
\end{itemize}
\

This kind of parallelization is useful not only to improve timing performance but also in scenarios in which map updates may occur from separated sensors in separated chunks, for example in multi-agent localization and mapping \cite{Parker2013}. For the record, parallelization across \emph{yNode}s is also possible but may lead to a computational overhead due to only  \emph{xNode}s being extensive.  Nonetheless, we plan to investigate on a possible deeper parallelization of the computation.

 \subsection{NEAREST NEIGHBOR AND RADIUS SEARCH}
 
 In a \emph{SkipList} the Nearest Neighbor Search is straightforward: when we search for a \emph{Key} in the list we always know the previous and next \emph{Key}s present in the set, even when the searching \emph{Key} is missing. A Radius Search around a target index is performed collecting all the elements between two indexes $I_{r^-},I_{r^+}$ obtained  starting from a center index and computing the boundaries with discrete radius dimension:
 
 \begin{equation}
	 I_{r^+},I_{r^-} = I \pm \floor*{\frac{radius}{resolution}}
 \end{equation}
 
\noindent as a \emph{SkipList} is an ordered linked-list, iterating from $I_{r^-}$ to $I_{r^+}$ allows for executing the search with  $\mathcal{O}(k)$ time complexity, $k$ being the number of elements within the range (Range Search). We can extend this approach to each of the \emph{SkipList}s present in our Tree, so to perform a Range Search along each $x,y,z$ dimension and obtain a Box Search. Then, filtering all the points found within the Box based on the distance from the box center allows for fetching a Sphere and thus achieve a Radius Search. As it will be shown in Section \ref{sec:results}, thanks to the  parallelization approach enabled by our data structure and discussed in previous section, our method outperforms standard implementation of search algorithms such as the \emph{KD-Tree} or \emph{Octree}.

\subsection{MAP UPDATE ON POSE GRAPH OPTIMIZATION}\label{section_integration}

\begin{figure}[thpb]
      \centering
      \includegraphics[width=0.35\textwidth]{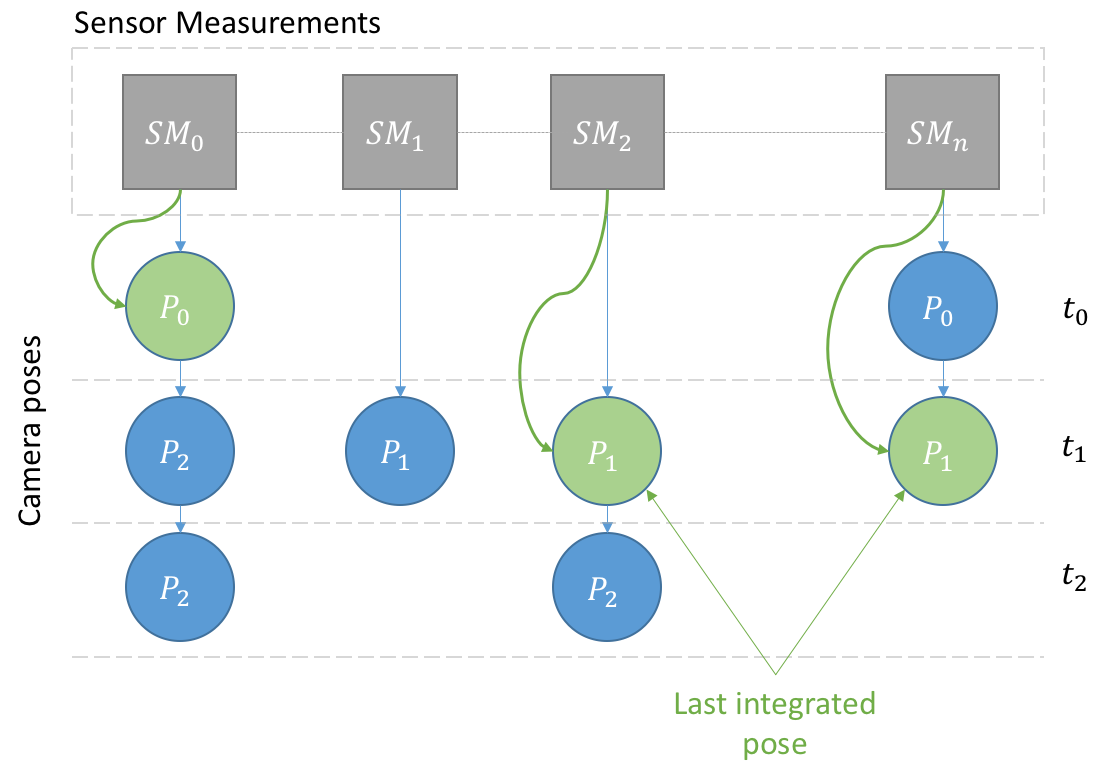}
      \caption{The Pose History consists of a set of queues associated with Sensor Measurements (SM). This structure allows for linking diffent poses to any SM so to keep track of which pose has been used to integrate them into the map as well as of the existence of newer ones possibly produced by the on-line pose optimization process. For example, at time $t_2$ the history linked to $SM_0$ shows that the mesaurements have been fused into the map according to $P_0$ but there exists a newer pose, i.e. $P_2$: the Pose Integrator may choose to erode $SM_0$ from the map according to $P_0$ and fuse measurements back according to $P_2$,  marking then the latter as the \emph{last integrated pose} for $SM_0$. Conversely, the last pose and last fused pose associated with $SM_n$ do coincide, so no action would be taken by Pose Integrator fot those measurements.}
      \label{img_pose_history}
   \end{figure}

The idea of \emph{Erosion} of past sensor measurements and \emph{Fusion} (or \emph{Integration}) of new ones  in a voxel grid was first introduced by Fioraio \emph{et. al.} \cite{Fioraio20152}. The integration procedure, described by Curless \emph{et. al.} \cite{curless1996}, allows to \emph{fuse} sensor measurement in a voxel grid according to a \emph{weight}; for example, to integrate the occupancy probability:

\begin{equation}\label{eq:fusion}
P'(v) = \frac{P(v)W(v)+p_i(v)w_i(i)}{W(v)+w_i(v)} , W'(v) = W(v) + w_i(v)
\end{equation}

\noindent where $P'(v),P(v)$ are the new and old occupancy probability of voxel $v$, respectively. $W'(v),W(v)$ the new and the old weight. As proposed in \cite{Fioraio20152}, the \emph{Erosion} process consists in just inverting the integration process:

\begin{equation}\label{eq:erosion}
P'(v) = \frac{P(v)W(v)-p_i(v)w_i(i)}{W(v)-w_i(v)} , W'(v) = W(v) - w_i(v)
\end{equation}

\noindent Erosion and fusion of sensor measurements may be deployed in conjunction with any sensor localization module capable of delivering optimized poses, e.g. upon detection of a loop closure. Thereby, the map may be updated by removing sensor measurements according to old poses and fusing them back according to the new, optimized poses. Our mapping system supports this feature by a \emph{weight} field and a generic data type associated with each  voxel, which allows the user to handle any desired kind of measurement (Occupancy Probability, SDF, RGB ....) in order to implement equations \ref{eq:fusion} and \ref{eq:erosion}.\par

However, though a sensor tracker typically produces poses at a certain controlled and approximately fixed pace (e.g. at every new sensor measurement or a controlled subset of them), optimized poses are delivered asynchronously with respect to such a regular rhythm, e.g. because a loop closure has been detected, and may happen to compete with live tracking as concerns updating the map. Therefore, as illustrated in Figure \ref{img_pose_history}, we have endowed \algoname{} with a \emph{Pose Manager} capable to create a \emph{Pose History}: the system treats live poses and optimized poses seamlessly  by inserting them in a set of queues, each associated with the sensor measurements (e.g. a depth image) taken at a certain time stamp; a \emph{Pose Integrator} chooses from the \emph{Pose History} a subset of poses and integrates the associated sensor measurements in the voxel map; if the pose that's about to be integrated is an optimized one, its predecessor will be \emph{eroded} from the map first. The choice of the subset of poses to be integrated into the map occurs according to the following criteria:

\begin{itemize}
\item live poses must be integrated as soon as possible.
\item among optimized poses, those spatially closer to the current live pose  are picked first.
\item the upper bound of the subset cardinality is fixed to ensure predictable computation time.
\end{itemize}

\subsection{GROUND TRACKING AND 2D QUERYING }\label{section_2d_query}

Although our proposal may be considered a generic 3D Mapping framework, it has been conceived to address robot navigation scenarios. Therefore, we found it useful to endow the framework with a module dedicated to tracking the ground plane. Thus, upon activation of the ground tracking module, the camera mounted on-board the robot must get a shot of the ground plane in the very first frame. The main plane found in the first frame is treated as the ground plane, which allows for classifying easily all the 3D points sensed in the successive frames as either \emph{ground} or \emph{obstacle} points.  This technique permits also to set the  \emph{Zero Reference Frame} of our map in the centroid of the first floor, thereby ensuring that $z$ coordinates are zero near the ground. More generally, if the core \algoname{} algorithm may be provided with measured points classified as \emph{ground} or \emph{obstacles}, they can be integrated in the map differently, and, in particular, so as to reduce the time complexity to integrate the former. Indeed,  with reference to Figure \ref{skiplist_volume}, integrating a \emph{ground point} boils down to allocating or updating only a \emph{green tile}/\emph{yNode} rather than a voxel, which implies reaching just depth level 2 of our SkipList Tree, whilst integrating \emph{obstacle} points would require going deeper to reach  level 3. \par
We can make the same point as visiting through the SkipList Tree: should we wish to retrieve only the information about ground in order to obtain a 2D Map we would need to visit the tree only up to depth level 2, thereby reducing  time complexity dramatically as vouched by Figure \ref{fig:graphs_visiting2d} . The figure shows also that the ability to create extremely rapidly a 2D view of the 3D Map is peculiar  to \algoname,  a classical approach like the \emph{Octree} being much slower due to the need to visit all voxels and project them on the ground in order to retrieve a 2D Map.

\subsection{IMPLEMENTATION DETAILS}

\algoname{} is implemented in C++ and wrapped in a ROS package, so to maximize its portability and usability in the robotics community. Thanks to widespread use of \emph{C++ Generics}, the \algoname{} data structure is contained in a couple of header files. Furthermore \emph{C++ Generics} enable to chose \emph{Data Type} to represent coordinates: for example, in our current implementation we have chosen \emph{short} as index data type allowing values in range $[-32.768,32.768]$, which results in a map of $655.36m$ along each dimension with a resolution of $0.01m$. Also voxels are \emph{templetized} so to allow the user to store whatever information therein. 

\section{RESULTS }
\label{sec:results}

\begin{figure*}[h]
  \centering
  \includegraphics[width=0.65\textwidth]{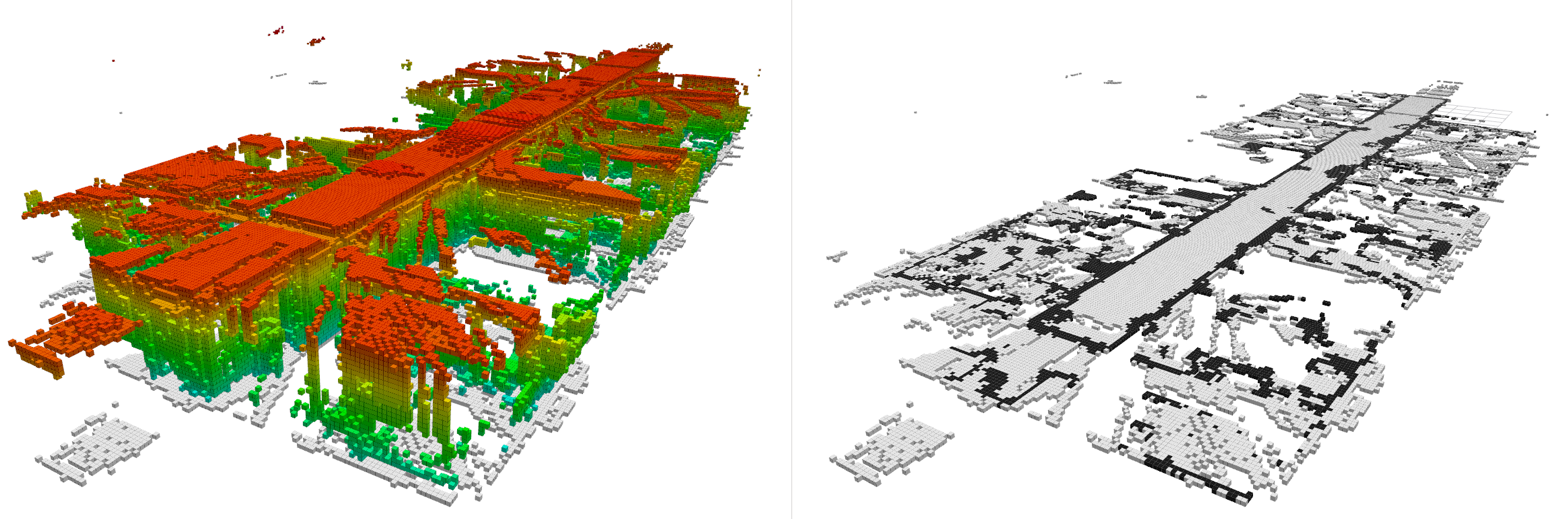}
  \caption{A Map built from Corridor Dataset collected in \emph{Octomap}\cite{Hornung2013}. \algoname{} allows for efficiently detecting the ground and, without further computational cost, discard higher obstacles  like the roof (red voxels in the left image) and labeling the ground voxels as \emph{navigable} (white regions in the right image).  }
  \label{fig:corridor}
  \vspace{0.05cm}
  \centering
  \includegraphics[width=0.65\textwidth]{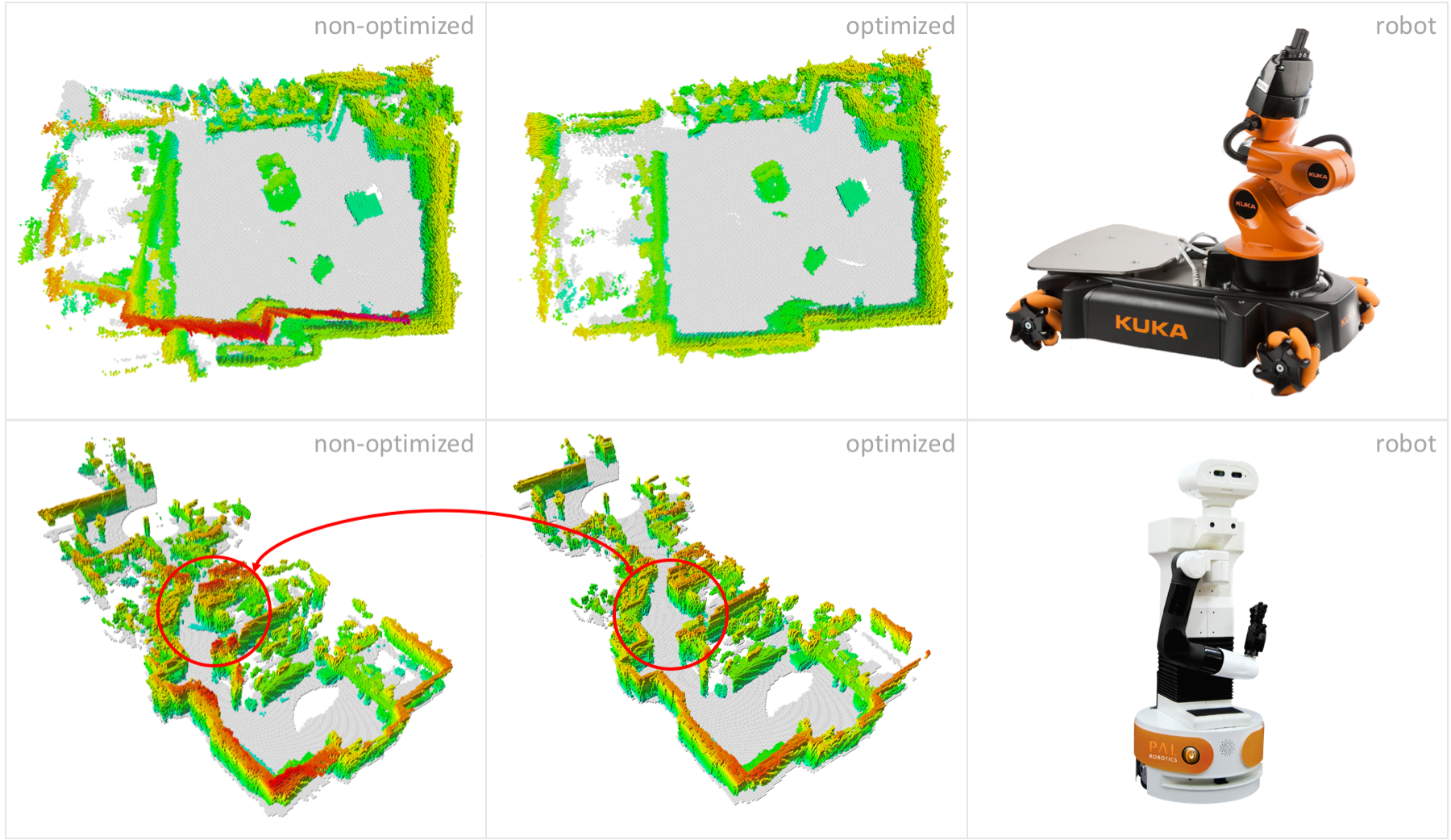}
  \caption{The first row concerns a small room ($5m\times4m\times3m$) reconstructed by \emph{Youbot} in \emph{eye-on-hand} configuration. The second row represents a medium-size environment ($8m\times35m\times3m$) reconstructed by \emph{Tiago} through an RGB-D camera mounted on the head. The middle column highlights the significant improvement in reconstruction accuracy provided by the real-time map optimization process.}
  \label{fig:tiago}
\end{figure*}

The \algoname{} mapping framework has been evaluated using some heterogeneous datasets categorized as follows:

\begin{itemize}
\item Medium-sized datasets captured with RGB-D sensors \cite{sturm2012}.
\item Public large-sized datasets captured with laser scanners mounted on pan-tilt units (\emph{Freiburg Campus}  \footnote{Courtesy of B. Steder, available at \url{
http://ais.informatik.uni-freiburg.de/projects/datasets/fr360/}}, \emph{New College} \cite{smith2009}).
\item Small and Medium-sized datasets captured in our Lab through RGB-D sensor on mobile robots (Figure \ref{fig:tiago}).
\end{itemize}

\begin{figure*}[h]
  \centering
  \includegraphics[width=\textwidth]{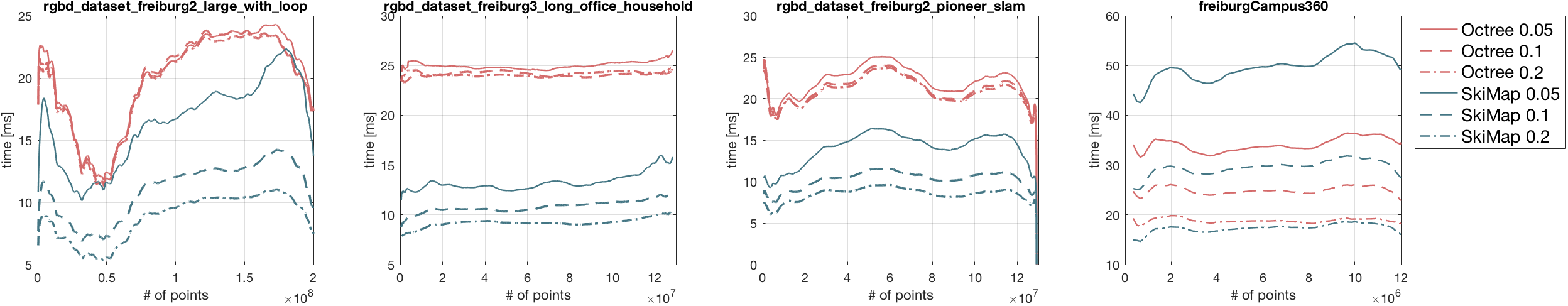}
  \caption{Time to integrate new measurements into the map with increasing number of total points. The first three datasets deal with RGB-D sensors ($\sim$ 320k points per scan) while the last one was acquired by a Laser Scanner mounted on Pan-Tilt unit ($\sim$ 180k points per scan). SkiMap provides inferior performance in the last dataset due to the scans featuring very spread and distant points (up to $50m$).}
  \label{fig:graphs_integration}
	
   \vspace{0.15cm}
 
  \centering
  \includegraphics[width=\textwidth]{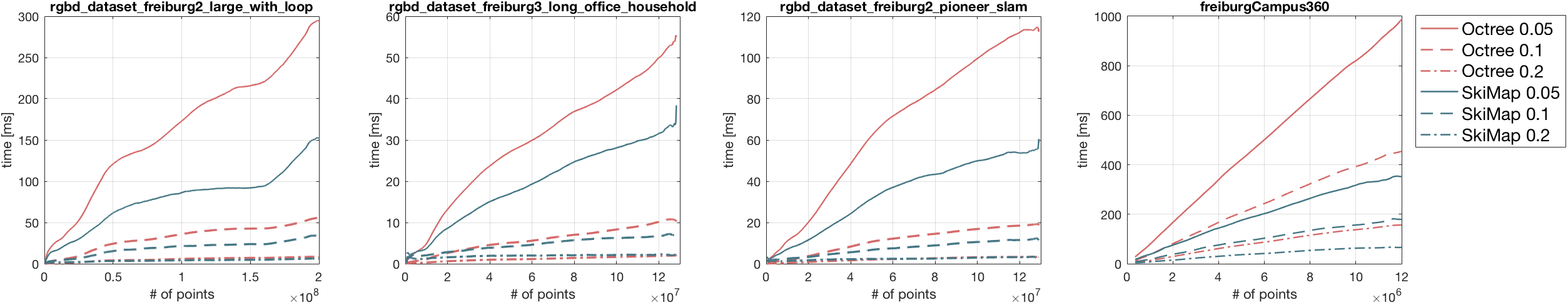}
  \caption{Time to visit the whole map.  }
  \label{fig:graphs_visiting}
  \vspace{0.15cm}

  \centering
  \includegraphics[width=\textwidth]{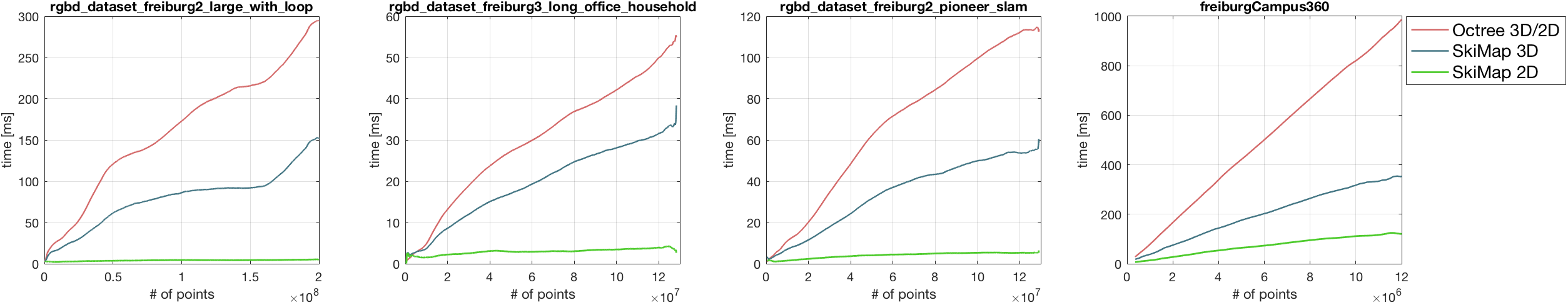}
  \caption{Comparison between 3D and 2D reconstructions. The \emph{Octree} requires the same time to perform  a full 3D or a 2D reconstruction because in both cases it needs to iterate over all the 3D points.  \algoname{}, instead, turns out faster than the \emph{Octree} in obtaining a 3D map as well as much faster in creating a 2D map thanks to the \emph{2D Query} feature. }
  \label{fig:graphs_visiting2d}
 	\vspace{0.15cm}
 	
 	\centering
  \includegraphics[width=\textwidth]{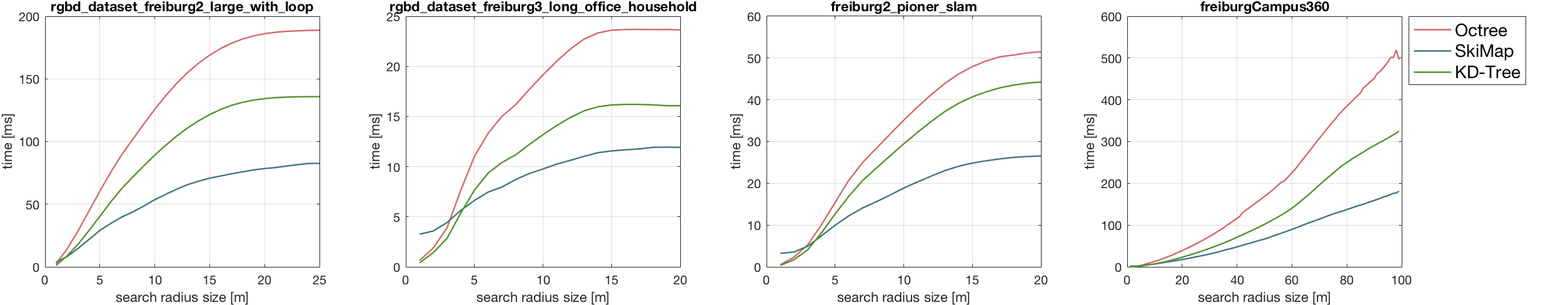}
  \caption{ Time to perform a radius search with increasing of radius size. \algoname{} outperforms  both the \emph{Octree} and the \emph{kd-tree} on all datasets.  }
  \label{fig:graphs_radius}
  
\end{figure*}

\noindent The public datasets are endowed with ground truth camera poses, while in the experiments concerning our datasets we deploy  \emph{Slamdunk}\cite{Fioraio2015} to track the camera. Thus, 
the quantitative evaluation reported in Figures \ref{fig:graphs_integration}, \ref{fig:graphs_visiting}, \ref{fig:graphs_visiting2d}, \ref{fig:graphs_radius} deals with the first two categories only - because of the availability of ground poses - and concerns a comparison between \algoname{} and the \emph{Octree}\footnote{version used: \url{https://github.com/OctoMap/octomap}} that is,  to the best of our knowledge, the foremost mapping solution in terms of memory efficiency. To attain a more comprehensive assessment, for each dataset we have considered multiple map resolutions, i.e. $0.05m$, $0.1m$ and $0.2m$. In Figure \ref{fig:graphs_radius} we have considered also the \emph{kd-tree}\footnote{version used: \url{http://pointclouds.org/}} because of its wide adoption in spatial search tasks such as \emph{radius search}. All the experiments have been run on a 5th generation Intel Core i7.

First we have assessed  basic tasks like \emph{``Integrating New Measurements''} (Figure \ref{fig:graphs_integration}) and \emph{``Visiting the Map"} (Figure \ref{fig:graphs_visiting}), finding out that \algoname{} is almost always more efficient than the \emph{Octree}. Figure \ref{fig:graphs_visiting2d} highlights  how the \emph{2D Query} feature introduced in \ref{section_2d_query} enables to outperform the \emph{Octree} in obtaining a similar representation. A qualitative example of the \emph{2D Query} feature can also be seen in Figure \ref{fig:corridor}, with the ground correctly reconstructed; it is worthwhile pointing out here that, as vouched by Figure \ref{fig:graphs_visiting2d}, obtaining this kind of representation by performing per-voxel projection to ground would imply a significantly higher time complexity. Finally, Figure \ref{fig:graphs_radius} is about the timing performance of the \emph{radius search} task, quite relevant, e.g., for the sake of avoiding obstacles while navigating within the workspace under reconstruction  Figure \ref{fig:graphs_radius} points out the much higher efficiency of \algoname{} with respect to both \emph{Octree} and \emph{kd-tree}, even without considering the initialization time to build the index required by the \emph{kd-tree} which is not accounted for in the Figure. As for memory occupancy, Table \ref{table:memory_footprint} highlights how \algoname{} tend to be almost as efficient as the \emph{Octree} in case of large environments while providing less memory savings with smaller workspaces.   

As for the experiments dealing with datasets taken in our Lab, we used  two mobile robots, namely \emph{Youbot} \cite{kuka2011} and \emph{Tiago} \footnote{\url{http://tiago.pal-robotics.com/}}, equipped with a Asus Xtion RGB-D sensor and, rather than relying on ground truth information, deployed \emph{SlamDunk}\cite{Fioraio2015} to track the robot/camera 6-DOF pose and  fuse sensor measurements into the map according to the estimated poses. Furthermore, leveraging on the \emph{Pose Optimization} module offered  by SlamDunk, we can realize the \emph{Map Update} feature of \algoname{} (see Section \ref{section_integration}). Both robots were operated manually, in small (\emph{Youbot}) and medium (\emph{Tiago}) sized  environments within our Lab, so to collect and fuse together multiple sensor measurements in order to reconstruct a map of the explored workspace. Figure \ref{fig:tiago} depicts examples of reconstructed maps with and without the \emph{Map Update} process enabled  by  SlamDunk's \emph{Pose Optimization} module. It is worthwhile pointing out that with our approach the optimized maps are not attained off-line within a post-processing step but built in real-time as described in \ref{section_integration}.



\begin{table}[]
\vspace{0.25cm}
\centering
\resizebox{0.35\textwidth}{!}{
\begin{tabular}{|l|l|l|l|l|}
\hline
\rowcolor[HTML]{C0C0C0} 
\textbf{Dataset}                                                                                         & \textbf{Type}      & \multicolumn{3}{l|}{\cellcolor[HTML]{C0C0C0}\textbf{Memory Saving wrt 3D Grid}}                                        \\ \hline
                                                                                                         &                         & \multicolumn{3}{l|}{\cellcolor[HTML]{C0C0C0}Resolutions}                                                               \\ \cline{3-5} 
                                                                                                         &                         & \cellcolor[HTML]{C0C0C0}\textbf{0.05m} & \cellcolor[HTML]{C0C0C0}\textbf{0.1m} & \cellcolor[HTML]{C0C0C0}\textbf{0.2m} \\ \hline
\rowcolor[HTML]{EFEFEF} 
\textbf{\begin{tabular}[c]{@{}l@{}}Freiburg Campus\\ $292\times167\times28 m^3$\end{tabular}}                            &                         &                                        &                                       &                                       \\ \hline
                                                                                                         & octree                  & 98.75\%                                & 96.21\%                               & 90.61\%                               \\
                                                                                                         & skimap                  & 98.52\%                                & 94.28\%                               & 83.33\%                               \\ \hline
\rowcolor[HTML]{EFEFEF} 
{\color[HTML]{000000} \textbf{\begin{tabular}[c]{@{}l@{}}New Dataset College\\ $250\times161\times33 m^3$\end{tabular}}} & {\color[HTML]{000000} } & {\color[HTML]{000000} }                & {\color[HTML]{000000} }               & {\color[HTML]{000000} }               \\ \hline
                                                                                                         & octree                  & 99.74\%                                & 99.00\%                               & 96.76\%                               \\
                                                                                                         & skimap                  & 99.77\%                                & 98.84\%                               & 95.30\%                               \\ \hline
\rowcolor[HTML]{EFEFEF} 
\textbf{\begin{tabular}[c]{@{}l@{}}Freiburg Long Office\\ $23\times25\times10 m^3$\end{tabular}}                         &                         &                                        &                                       &                                       \\ \hline
                                                                                                         & octree                  & 90.50\%                                & 84.71\%                               & 74.91\%                               \\
                                                                                                         & skimap                  & 82.62\%                                & 71.30\%                               & 54.63\%                               \\ \hline
\end{tabular}
}

\caption{Percentage of memory savings with respect to a full 3D grid.}

\label{table:memory_footprint}

\end{table}


\section{CONCLUDING REMARKS}

In this work we have described a novel mapping approach  mainly devoted to robot navigation. The primary objective was to provide an efficient mapping framework suitable to real-time applications in embedded robotics platforms. Thus, unlike approaches that focus on dense and accurate 3D reconstruction, such as e.g. \cite{Dai2016}, our method is aimed at building as efficiently as possible the kinds of representation required to support robot navigation effectively. In its current state the framework can also provide some basic form of semantic information, such as telling apart \emph{ground} and \emph{obstacles}. We plan to enrich the degree of semantic perception accommodated by \algoname{} by incorporating  detection of certain object instances \cite{Fioraio2013CVPR}, e.g. items to be picked or manipulated by the robot, as well as by leveraging on per-frame \emph{Semantic Segmentation} so to fuse \emph{category labels} into the map \cite{cavallari2015}, \cite{Cavallari2016}.




\bibliographystyle{IEEEtran}
\bibliography{IEEEabrv,bibliography}

\begin{thebibliography}{10}
\providecommand{\url}[1]{#1}
\csname url@rmstyle\endcsname
\providecommand{\newblock}{\relax}
\providecommand{\bibinfo}[2]{#2}
\providecommand\BIBentrySTDinterwordspacing{\spaceskip=0pt\relax}
\providecommand\BIBentryALTinterwordstretchfactor{4}
\providecommand\BIBentryALTinterwordspacing{\spaceskip=\fontdimen2\font plus
\BIBentryALTinterwordstretchfactor\fontdimen3\font minus
  \fontdimen4\font\relax}
\providecommand\BIBforeignlanguage[2]{{%
\expandafter\ifx\csname l@#1\endcsname\relax
\typeout{** WARNING: IEEEtran.bst: No hyphenation pattern has been}%
\typeout{** loaded for the language `#1'. Using the pattern for}%
\typeout{** the default language instead.}%
\else
\language=\csname l@#1\endcsname
\fi
#2}}

\bibitem{ROS09}
M.~Quigley, K.~Conley, B.~P. Gerkey, J.~Faust, T.~Foote, J.~Leibs, R.~Wheeler,
  and A.~Y. Ng, ``Ros: an open-source robot operating system,'' in \emph{ICRA
  Workshop on Open Source Software}, 2009.

\bibitem{Fioraio2015}
N.~Fioraio and L.~Di~Stefano, ``Slamdunk: Affordable real-time rgb-d slam,'' in
  \emph{Computer Vision - ECCV 2014 Workshops: Zurich, Switzerland, September
  6-7 and 12, 2014, Proceedings, Part I}, 2015, pp. 401--414.

\bibitem{Thrun05}
S.~Thrun, W.~Burgard, and D.~Fox, \emph{Probabilistic Robotics (Intelligent
  Robotics and Autonomous Agents)}.\hskip 1em plus 0.5em minus 0.4em\relax The
  MIT Press, 2005.

\bibitem{Grisetti07}
G.~Grisetti, C.~Stachniss, and W.~Burgard, ``Improved techniques for grid
  mapping with rao-blackwellized particle filters,'' \emph{IEEE Transactions on
  Robotics}, vol.~23, no.~1, pp. 34--46, Feb 2007.

\bibitem{Tabak89}
Y.~Roth-Tabak and R.~Jain, ``Building an environment model using depth
  information,'' \emph{Computer}, vol.~22, no.~6, pp. 85--90, June 1989.

\bibitem{meagher82}
D.~J. Meagher, ``Geometric modeling using octree encoding,'' in \emph{Computer
  Graphics and Image Processing}, 1982, pp. 129--147.

\bibitem{Hornung2013}
\BIBentryALTinterwordspacing
A.~Hornung, K.~M. Wurm, M.~Bennewitz, C.~Stachniss, and W.~Burgard, ``Octomap:
  an efficient probabilistic 3d mapping framework based on octrees,''
  \emph{Autonomous Robots}, vol.~34, no.~3, pp. 189--206, 2013. [Online].
  Available: \url{http://dx.doi.org/10.1007/s10514-012-9321-0}
\BIBentrySTDinterwordspacing

\bibitem{Labschutz16}
M.~Labschütz, S.~Bruckner, M.~E. Gröller, M.~Hadwiger, and P.~Rautek,
  ``Jittree: A just-in-time compiled sparse gpu volume data structure,''
  \emph{IEEE Transactions on Visualization and Computer Graphics}, vol.~22,
  no.~1, pp. 1025--1034, Jan 2016.

\bibitem{Triebel06}
R.~Triebel, P.~Pfaff, and W.~Burgard, ``Multi-level surface maps for outdoor
  terrain mapping and loop closing,'' in \emph{2006 IEEE/RSJ International
  Conference on Intelligent Robots and Systems}, Oct 2006, pp. 2276--2282.

\bibitem{dryanovski10}
I.~Dryanovski, W.~Morris, and J.~Xiao, ``Multi-volume occupancy grids: An
  efficient probabilistic 3d mapping model for micro aerial vehicles,'' in
  \emph{Intelligent Robots and Systems (IROS), 2010 IEEE/RSJ International
  Conference on}, Oct 2010, pp. 1553--1559.

\bibitem{Gutmann05}
J.~S. Gutmann, M.~Fukuchi, and M.~Fujita, ``A floor and obstacle height map for
  3d navigation of a humanoid robot,'' in \emph{Proceedings of the 2005 IEEE
  International Conference on Robotics and Automation}, April 2005, pp.
  1066--1071.

\bibitem{Maier12}
D.~Maier, A.~Hornung, and M.~Bennewitz, ``Real-time navigation in 3d
  environments based on depth camera data,'' in \emph{2012 12th IEEE-RAS
  International Conference on Humanoid Robots (Humanoids 2012)}, Nov 2012, pp.
  692--697.

\bibitem{Biswas12}
J.~Biswas and M.~Veloso, ``Depth camera based indoor mobile robot localization
  and navigation,'' in \emph{Robotics and Automation (ICRA), 2012 IEEE
  International Conference on}, May 2012, pp. 1697--1702.

\bibitem{Fioraio20152}
N.~Fioraio, J.~Taylor, A.~Fitzgibbon, L.~D. Stefano, and S.~Izadi,
  ``Large-scale and drift-free surface reconstruction using online subvolume
  registration,'' in \emph{2015 IEEE Conference on Computer Vision and Pattern
  Recognition (CVPR)}, June 2015, pp. 4475--4483.

\bibitem{Dai2016}
A.~Dai, M.~Nie{\ss}ner, M.~Zollh\"ofer, S.~Izadi, and C.~Theobalt,
  ``Bundlefusion: Real-time globally consistent 3d reconstruction using online
  surface re-integratio,'' \emph{arXiv preprint arXiv:1604.01093}, 2016.

\bibitem{niessner2013hashing}
M.~Nie{\ss}ner, M.~Zollh\"ofer, S.~Izadi, and M.~Stamminger, ``Real-time 3d
  reconstruction at scale using voxel hashing,'' \emph{ACM Transactions on
  Graphics (TOG)}, 2013.

\bibitem{Pugh90}
W.~Pugh, ``Skip lists: A probabilistic alternative to balanced trees,''
  \emph{Commun. ACM}, pp. 668--676, 1990.

\bibitem{Pugh89C}
------, ``Concurrent maintenance of skip lists,'' Tech. Rep., 1998.

\bibitem{Parker2013}
L.~E. Parker, K.~Fregene, Y.~Guo, and R.~Madhavan, ``Multi-robot localization,
  mapping, and path planning,'' in \emph{Multi-Robot Systems: From Swarms to
  Intelligent Automata: Proceedings from the 2002 NRL Workshop on Multi-Robot
  Systems}.\hskip 1em plus 0.5em minus 0.4em\relax Springer Science \& Business
  Media, 2013, p.~21.

\bibitem{curless1996}
B.~Curless and M.~Levoy, ``A volumetric method for building complex models from
  range images,'' in \emph{Proceedings of the 23rd annual conference on
  Computer graphics and interactive techniques}.\hskip 1em plus 0.5em minus
  0.4em\relax ACM, 1996, pp. 303--312.

\bibitem{sturm2012}
J.~Sturm, N.~Engelhard, F.~Endres, W.~Burgard, and D.~Cremers, ``A benchmark
  for the evaluation of rgb-d slam systems,'' in \emph{2012 IEEE/RSJ
  International Conference on Intelligent Robots and Systems}.\hskip 1em plus
  0.5em minus 0.4em\relax IEEE, 2012, pp. 573--580.

\bibitem{smith2009}
\BIBentryALTinterwordspacing
M.~Smith, I.~Baldwin, W.~Churchill, R.~Paul, and P.~Newman, ``{The new college
  vision and laser data set},'' \emph{The International Journal of Robotics
  Research}, vol.~28, no.~5, pp. 595--599, May 2009. [Online]. Available:
  \url{http://www.robots.ox.ac.uk/NewCollegeData/}
\BIBentrySTDinterwordspacing

\bibitem{kuka2011}
R.~Bischoff, U.~Huggenberger, and E.~Prassler, ``Kuka youbot-a mobile
  manipulator for research and education,'' in \emph{Robotics and Automation
  (ICRA), 2011 IEEE International Conference on}.\hskip 1em plus 0.5em minus
  0.4em\relax IEEE, 2011, pp. 1--4.

\bibitem{Fioraio2013CVPR}
N.~Fioraio and L.~Di~Stefano, ``Joint detection, tracking and mapping by
  semantic bundle adjustment,'' in \emph{The IEEE Conference on Computer Vision
  and Pattern Recognition (CVPR)}, June 2013.

\bibitem{cavallari2015}
T.~Cavallari and L.~Di~Stefano, ``Volume-based semantic labeling with signed
  distance functions,'' in \emph{Pacific-Rim Symposium on Image and Video
  Technology}.\hskip 1em plus 0.5em minus 0.4em\relax Springer International
  Publishing, 2015, pp. 544--556.

\bibitem{Cavallari2016}
------, ``On-line large scale semantic fusion,'' in \emph{Computer Vision --
  ECCV 2016 Workshops: Amsterdam, The Netherlands, October 8-10 and 15-16,
  2016, Proceedings, Part III}, 2016, pp. 83--99.

\end{thebibliography}

\end{document}